\documentclass[10pt, a4paper]{article}
\usepackage{lrec2022} 
\usepackage{multibib}
\newcites{languageresource}{Language Resources}
\usepackage{graphicx}
\usepackage{tabularx}
\usepackage{soul}
\usepackage{titlesec}
\titleformat{\section}{\normalfont\large\bfseries\center}{\thesection.}{1em}{}
\titleformat{\subsection}{\normalfont\SmallTitleFont\bfseries\raggedright}{\thesubsection.}{1em}{}
\titleformat{\subsubsection}{\normalfont\normalsize\bfseries\raggedright}{\thesubsubsection.}{1em}{}
\renewcommand\thesection{\arabic{section}}
\renewcommand\thesubsection{\thesection.\arabic{subsection}}
\renewcommand\thesubsubsection{\thesubsection.\arabic{subsubsection}}

\usepackage{epstopdf}
\usepackage[utf8]{inputenc}

\usepackage{hyperref}
\usepackage{xstring}

\usepackage{color}
\usepackage{multirow}

\title{Annotation Errors and NER: A Study with OntoNotes 5.0}

\name{Gabriel Bernier-Colborne, Sowmya Vajjala} 

\address{National Research Council Canada \\
         {\tt \{Gabriel.Bernier-Colborne | Sowmya.Vajjala\}@nrc-cnrc.gc.ca}
         }
\abstract{
Named Entity Recognition (NER) is a well-studied problem in NLP. However, there is much less focus on studying NER datasets, compared to developing new NER models. In this paper, we employed three simple techniques to detect annotation errors in the OntoNotes 5.0 corpus for English NER, which is the largest available NER corpus for English. Our techniques corrected $\sim 10\%$ of the sentences in train/dev/test data. In terms of entity mentions, we corrected the span and/or type of $\sim 8\%$ of mentions in the dataset, while adding/deleting/splitting/merging a few more. These are large numbers of changes, considering the size of OntoNotes. We used three NER libraries to train, evaluate and compare the models trained with the original and the re-annotated datasets, which showed an average improvement of $1.23\%$ in overall F-scores, with large ($>10$\%) improvements for some of the entity types. While our annotation error detection methods are not exhaustive and there is some manual annotation effort involved, they are largely language agnostic and can be employed with other NER datasets, and other sequence labelling tasks.
 \\ \newline }

\begin{document}
\maketitleabstract

\section{Introduction}
\label{sec:intro}
NLP relies on the presence of high-quality, large-scale annotated corpora for the development and evaluation of various models and their applications. Such corpora are generally created with a lot of human effort and expertise along with some automatic or semi-automatic means. Some amount of annotation errors seems unavoidable in this process, and they can pose challenges for reliable training and evaluation of NLP systems. Identifying and correcting such errors is important to understand model performance and limitations for any task. Past NLP research has explored methods to identify such errors through various approaches (see \newcite{dickinson2015detection} for a broad survey) from re-annotation studies to using predictions from systems trained on a given annotated dataset. 

Named Entity Recognition (NER) is a well-studied research problem in NLP, and is also widely used in applied contexts. State-of-the-art NER systems report high performance ($>$ 90\% F-score) for English, on commonly used datasets.\footnote{\url{http://nlpprogress.com/english/named_entity_recognition.html}} NER is traditionally treated as a supervised learning problem, and hence relies on the presence of large, annotated datasets. However, NER modelling and evaluation is usually carried out by considering the dataset as an unalterable gold standard, and there is little research that looks into the quality of the datasets themselves. Existing research on identifying and correcting annotation errors in NER primarily focuses on the CONLL-03 dataset~\cite{sang2003introduction}. The NER component of OntoNotes 5.0~\cite{weischedel2013ontonotes} is the largest available NER dataset for English, but we are not aware of any studies on this dataset. 

In this paper, we: 
\begin{enumerate}
    \item Detect and correct many annotation errors in OntoNotes 5.0 English NER dataset. Our methods involving data analysis followed by manual re-annotation can be applied on any NER dataset, in any language. 
    \item Demonstrate the effect of the observed annotated errors on NER model performance by comparing three NER systems with the original and re-annotated datasets, across different entity types. 
\end{enumerate}

Our approach corrected close to 10\% of the sentences in the dataset. To our knowledge, this is the first paper to identify annotation errors in OntoNotes 5.0 English NER dataset, and study their impact on NER performance. 

The remaining sections of this paper are organized as follows: We give an overview of related literature in Section~\ref{sec:relw} and describe our methodology for annotation error detection in Section~\ref{sec:methods}. The results of our manual corrections are discussed in Section~\ref{sec:results1} and the impact of this work on model performance is evaluated in Section~\ref{sec:results2}. Section~\ref{sec:concl} discusses the main conclusions of this paper.  

\section{Related Work}
\label{sec:relw}
Detecting and correcting annotation errors, and assessing the quality of annotated corpora has been an area of interest in NLP research. For example,  \newcite{silberztein-2018-using} discussed ways to use existing linguistic resources to assess the quality of annotated corpora in terms of their linguistic coverage and identify potential annotation errors in POS tagging. Previous research also explored ways to detect annotation errors in dependency treebanks by looking for annotation differences between identical word sequences~\cite{wisniewski-2018-errator} and using a parse quality assessment algorithm~\cite{alzetta-etal-2017-dangerous}. Such efforts have primarily focused on POS tagging and dependency/constituency parsing corpora. In contrast to this research on detecting annotation errors on human-labelled data, \newcite{rehbein-ruppenhofer-2017-detecting} proposed an approach to detect such errors in automatically labelled data through active learning. 

Detecting annotation errors in NER is a recent topic of interest in NLP research. For the related task of entity linking, \newcite{jha2017all} proposed a rule-based approach to semi-automatically check a gold standard corpus for potential annotation errors, and evaluated with multiple datasets. A commonly employed approach for NER is to study the predictions of existing NER models on a standard dataset to identify and classify errors made by the models, and identify annotation errors as a part of this process. 
\newcite{stanislawek-etal-2019-named} developed a taxonomy of the errors made by state-of-the-art NER systems on the CONLL-03 dataset, which included some annotation-level errors. However, this analysis was done only on the test fold of the dataset. \newcite{reiss-etal-2020-identifying,muthuraman-etal-2021-data} used the predictions from multiple NER models and model ensembles to identify  annotation errors in the entire CONLL-03 English NER corpus and identified 3.76\% incorrect NE labels in the entire dataset, and 7.45\% in the test set alone.  

Research in this direction for other languages followed a similar procedure. \newcite{abudukelimu-etal-2018-error} used the predictions from a state-of-the-art NER tagger for the Uyghur language to identify and categorize annotation errors as well as other errors in the model predictions. \newcite{ichihara2015error} followed this approach on a Japanese NER corpus/tagger. \newcite{saha2009hindi} used a combination of rules and a feature-based NER tagger to identify and correct annotation errors in a Hindi dataset. 

Instead of building and evaluating NER models to identify errors in the data, \newcite{wang-etal-2019-crossweigh} re-annotated the English test set of CONLL-03 dataset and discovered labelling mistakes in 5.38\% of the sentences in the test set.  

In this paper, we neither rely on predictions from NER models to identify potential errors nor take the ``re-annotate the entire test set'' approach. Instead we rely on simple means of identifying potential mis-annotations and manually re-annotate them in all three parts of the corpus (i.e. training, development, and test). We then train and test NER models with both old and new datasets. To our knowledge, this is the first paper to address this issue with OntoNotes, the largest English NER dataset available as of now.     

\section{Methods}
\label{sec:methods}
Ontonotes 5.0 is a large dataset comprising multi-genre text in English, Arabic, and Chinese, with linguistic annotations covering multiple layers such as part of speech, parse structure, co-reference, proposition, word sense and NER. The English NER portion of this corpus consists of 3,637 documents, covering 2 million tokens, annotated with 18 named entity types~\cite{Pradhan2013TowardsRL}. It is among the most commonly used NER datasets for English in the past decade.\footnote{\url{http://nlpprogress.com/english/named_entity_recognition.html}}

OntoNotes' release notes~\cite{weischedel2013ontonotes} do not describe the annotation process in detail, but mentions that each layer of annotation has at least 90\% inter-annotator agreement. Regarding NER annotation specifically, though, \newcite{weischedel2011ontonotes} say: ``Names are also annotated using an 18-type superset of the ACE name guidelines. This supplemental annotation is done in a single pass.'' The document \textit{OntoNotes Named Entity Guidelines} (version 14.0), which is included in the LDC release and which we will refer to simply as ``the guidelines'' below, gives a rough outline of how to annotate each of the 18 entity types, with a few examples. These guidelines form the basis of our study in this paper, and we tried to follow them as closely as possible when re-annotating the dataset. 

Based on these guidelines, we categorized potential annotation errors into the following four groups:
\begin{enumerate}
    \item Deviations from the guidelines. For example, the guidelines specify that determiners or articles should not be included in the span (except for certain numerical entity types), suggesting to mark only ``US'' in ``the US'', and ``White House'' in ``the White House'' as examples. However, those two entities appear very frequently in all partitions of the dataset and marked with ``the'' included in many instances. 
    \item Inconsistent annotations, where the guidelines are not clear, and the annotations are not consistent throughout. Inclusion of possessive markers as a part of the entity is an example. They are included in some entity mentions, and excluded in others. For example, ``Boris Yelstin's'' is tagged PERSON in ``Boris Yelstin's successor'' but only ``Yasser Arafat'' is tagged in ``Yasser Arafat's proposal.''
    \item Wrong annotations, where the annotated NE tag is clearly wrong (e.g., a non-entity tagged as an entity, an entity tagged with the wrong type or span, etc.). For example, in the sentence ``Regarding the joint development he talked about, Russia is also very interesting'', ``interesting'' is tagged as a GPE. 
    \item Ambiguous annotations, where it is not clear whether the annotation is correct. For example, in the sentence `He describes a reporter as ``Miss First Amendment.''', ``First Amendment'' is tagged as LAW. Going by the guidelines, ``unique nicknames'' should be tagged a PERSON, which makes ``Miss First Amendment'' a PERSON. However, it is not clear from the guidelines if this is how it should be understood. 
\end{enumerate}

We corrected the first three forms of errors we identified through our approach (described below), and left the fourth category (ambiguous annotations) as is. We did not address issues related to tokenization or part of speech, or entities that may need more context beyond a sentence to disambiguate, although these exist too. 

\subsection{Our Approach}

We followed three methods to identify potential annotation errors and corrected them manually. These are described below.

\paragraph{Complete Manual Examination} OntoNotes is a large dataset consisting of 2 million tokens in total. Hence, complete manual re-examination could be a challenging endeavour. Hence, we chose 4 of the least frequent entity types (LANG, LAW, EVENT, PRODUCT) in the test set and manually examined the mentions for these in train/dev/test sets, correcting for the three groups of annotation errors described above.

\paragraph{HardEval} 

The first step of this method consists in inspecting potentially erroneous mentions out of context by looking for tokens or labels that are somehow ``suprising'', following the HardEval methodology~\cite{bernier2020hardeval}, which was originally designed to evaluate NER systems by focusing on challenging subsets of tokens. In this work, we looked at all unique mentions that contained either of three types of tokens: unseen tokens, tokens that are usually not part of a mention, or tokens that are usually part of a mention of a different entity type. These correspond to the ``unseen-I'', ``diff-I'', and ``diff-etype'' subsets of HardEval. These subsets of tokens are identified by comparing token or label frequencies in the dev or test set to those in the training set. To identify these token subsets in the training set, we used 10-fold cross validation. 

This produced over 20K unique mentions to check for the training set. All of the potential errors were inspected, out of context, and attention was paid specifically to unseen-I, diff-I or diff-etype tokens. Mentions that were indeed suspicious were then inspected in context and re-annotated if appropriate.

This process revealed some very frequent re-annotations that could be automated, in order to speed up the process, apply corrections systematically, and guard against introducing new errors. Therefore we developed a script that interactively applies conditional re-annotation rules. Several of the annotation guidelines can be expressed as such rules (though the precision/recall may not be perfect). Three such rules have been implemented so far: one for leading determiners, one for trailing possessive markers, and one for leading and/or trailing punctuation. For instance, in the case of leading determiners, it checks for mentions starting with ``the'' or ``a'', and that are neither DATE nor TIME (as these can and often do start with a determiner), and for each of these mentions, it asks the user whether the first token should be removed from the mention. Processing the several thousand mentions raised by the determiner rule over the entire training set took about an hour.

\paragraph{Selecting from Mention-Type lists}: We selected potential annotation errors by inspecting the most frequent $<$mention, type$>$ pairs for non-numeric entity types such as PERSON, GPE, LOC, FAC, and WORK\textunderscore{}OF\textunderscore{}ART. For example, the pair $<$first, PERSON$>$ appeared in the mention-type pairs list in the dev set, which was used to identify the sentence where the word ``first'' in ``for the first time'' was wrongly tagged B-PERSON instead of O. 

These three approaches are not exhaustive, and cannot find all annotation errors in the corpus. They also cannot find entity mentions that have been overlooked by the annotators (i.e. tagged as non-entities). Further, identifying and then manually correcting all the identified errors is manually intensive. However, the goal of this paper is primarily to highlight some of the annotation errors and their impact on the final NER model performance, rather than completely correcting all possible errors in an obviously large dataset, in a single go. 

\section{Re-Annotation}
\label{sec:results1}
Our re-annotation resulted in changes to close to 10\% of sentences in train/test/dev sets. This section shows some of the re-annotation statistics and discusses in detail our notes on the nature of changes, and some common problems we observed. 

\subsection{Re-annotation Statistics}

Table~\ref{tab:reannotation-stats} shows some statistics on the number of corrections made to the dataset at token/sentence/mention level. It distinguishes 7 different types of mention-level changes, i.e. deleted mentions, added mentions, mentions that were split into two or more, and mentions that were merged from two or more, as well as one-to-one alignments with either a span change or a type change, or both. These will be discussed in Section~\ref{sec:observations}.
\begin{table*}[ht]
    \centering
    \begin{tabular}{|l|l|l|l|}
        \hline
        Statistic & Train & Dev & Test \\
        \hline
		Nb changed tokens    & 14092 (1.29\%) & 1812 (1.23\%) & 1774 (1.16\%) \\
		Nb changed sentences & 6009 (10.03\%) & 747 (8.76\%)  & 777 (9.40\%)  \\
        \hline        
        Nb mentions before & 81828 & 11066 & 11257 \\
        Nb mentions after  & 81891 & 11041 & 11253 \\
        \hline
        Deleted mentions           & 320  & 22  & 21  \\
        Added mentions             & 353  & 3   & 3   \\
        Span changed, but not type & 6171 & 809 & 820 \\
        Type changed, but not span & 428  & 71  & 53  \\
        Both span and type changed & 118  & 12  & 19  \\
        Split into 2 or more       & 40   & 5   & 15  \\
        Merged from 2 or more      & 8    & 11  & 3   \\
        \hline
    \end{tabular}
	\caption{Statistics on corrections made to the dataset. The last 7 rows relate to different categories of mention-level changes.}
	\label{tab:reannotation-stats}
\end{table*}

As mentioned earlier, our approach of identifying and correcting annotation errors in OntoNotes resulted in correcting close to 10\% of sentences in each of train, dev and test sets (Table~\ref{tab:reannotation-stats}). This is a larger amount of corrections compared to 5.38\% of test sentences changed in the complete manual re-annotation of CONLL-03 test set by \newcite{wang-etal-2019-crossweigh}). In terms of the number of mentions, \newcite{reiss-etal-2020-identifying} report correcting 3.76\% of the mentions in the entire CONLL-03 dataset (7.45\% in the test data). Our approach corrected the span and/or type of 7.92\%/7.84\%/7.59\% of mentions in the original train/dev/test sets, while adding/deleting/splitting/merging a few more. Considering that OntoNotes is a much larger dataset than CONLL-03, the detected (and corrected) annotation errors identified are indeed substantial in number compared to previous work. 
Table~\ref{tab:reannotation-stats-by-type} shows the largest changes in entity type frequencies resulting from merging or splitting mentions, or correcting their entity type. Most changes happened for EVENT and WORK\_OF\_ART in the case of training and test sets (in terms of percentages), and for FAC and WORK\_OF\_ART in the development set. 
\begin{table*}[ht]
    \centering
    \begin{tabular}{|l|l|l|l|}
		\hline
		Entity type & Train & Dev & Test \\
		\hline
		EVENT       & -78 (-10.43\%) & -3 (-2.10\%)   & -12 (-19.05\%) \\
		FAC         & +26 (+3.02\%)  & +18 (+15.65\%) & +5 (+3.70\%) \\
		LANGUAGE    & +4 (+1.32\%)   & 0 (0.00\%)     & +1 (+4.55\%) \\
		LAW         & +5 (+1.77\%)   & +2 (+5.00\%)   & +1 (+2.50\%) \\
		LOC         & -67 (-4.43\%)  & -6 (-2.94\%)   & -8 (-4.47\%) \\
		ORG         & -49 (-0.38\%)  & -26 (-1.49\%)  & -23 (-1.28\%) \\
		PRODUCT     & -6 (-0.99\%)   & -1 (-1.39\%)   & +3 (+3.95\%) \\
		WORK\textunderscore{}OF\textunderscore{}ART & +57 (+5.85\%)  & +10 (+7.04\%)  & +28 (+16.87\%) \\
		\hline
	\end{tabular}
	\caption{Largest changes in entity type frequencies resulting from merging or splitting mentions, or correcting their entity type.}
	\label{tab:reannotation-stats-by-type}
\end{table*}

\subsection{Observations}
\label{sec:observations}
As shown in Table~\ref{tab:reannotation-stats}, our corrections included different types of mention-level changes. These are described in more detail below.

\textbf{Deleted mentions} include nominal mentions (e.g. ``the Minister'' or ``mom'' as PERSON, ``unions'' as ORG, ``territories'' as LOC, ``republics'' as GPE), e-mail addresses and URLs, expressions denoting the mafia, and words such as ``now'' labelled as TIME. They also include various expressions that are clearly not named entity mentions (e.g. ``in'' as GPE, ``light years'' as TIME, ``died with a smile on'' as CARDINAL, ``the fleas are gone'' as GPE).

\textbf{Added mentions} include any mention that we happened to find that was not labelled as such. We found overlooked mentions of various entity types, including: ``Putin'' (PERSON), ``Iraq''(GPE), ``National Palace Museum'' (FAC), ``CUNY'' (ORG), ``Conservatives'' (NORP), ``MBA'' (WORK\textunderscore{}OF\textunderscore{}ART), ``Arabic'' (LANGUAGE), ``Olympics'' (EVENT), etc. Although our methods are not designed to detect mentions that were overlooked, we did find a significant number of these as we were correcting other errors.

Examples of \textbf{split mentions} include the following: ``Lada Niva'' (PRODUCT) is split into ``Lada'' (ORG) and ``Niva'' (PRODUCT); ``Guo Kai’s Economic Notes'' (WORK\textunderscore{}OF\textunderscore{}ART) is split into ``Guo Kai'' (PERSON) and ``Economic Notes'' (WORK\textunderscore{}OF\textunderscore{}ART); ``the Supreme Court of the United States'' (ORG) is split into ``Supreme Court'' (ORG) and ``United States'' (GPE).

Examples of \textbf{merged mentions} include the following: ``Sunday'' (DATE) and ``evening'' (TIME) are merged into ``Sunday evening'' (TIME); ``Roe'' (PERSON) and ``Wade'' (PERSON) are merged into ``Roe v. Wade'' (LAW); ``Atlantic'' (LOC) and ``coast'' (LOC) are merged into ``Atlantic coast'' (LOC); and ``Deloitte'' (PERSON) and ``Haskins \& Sells'' (ORG) are merged into ``Deloitte, Haskins \& Sells'' (ORG).

\textbf{Span changes} are the most numerous category. The most frequent corrections involve the removal of leading articles or determiners (e.g. ``a'', ``an'', ``the'', ``this'', ``that''), trailing possessive markers, and leading or trailing punctuation. Corrections in this category also include the removal of: the leading words ``since'' or ``for'' from DATE mentions; the word ``language'' from LANGUAGE mentions (e.g. ``the Arabic language'' becomes ``Arabic''); and titles, honorifics or familial relation markers from PERSON mentions (e.g. ``Minister'', ``Dr.'', ``Uncle''). Also, the span of MONEY mentions is enlarged to include the monetary unit if it was excluded.

\textbf{Type changes} include a wide variety of examples:

\begin{itemize}
\item Some ORGs become FAC (e.g. names of airports or churches), WORK\textunderscore{}OF\textunderscore{}ART (e.g. TV show names) or DATE (e.g. names of dynasties).
\item Some PRODUCTs (e.g. ``BMW'') or WORK\textunderscore{}OF\textunderscore{}ARTs (e.g. names of newspapers or magazines) become ORG. 
\item Some LOCs become FAC (e.g. names of streets and parks) or GPE (e.g. ``NYC'', ``California'', ``Germany''), and some GPEs become LOC (e.g. ``Provence'', ``Africa'')
\item Some LOCs or GPEs become NORP (e.g. ``African'', ``Japanese''), whereas some NORPs become LANGUAGE (e.g. ``Cantonese''), GPE (e.g. ``Australia'') or ORG (e.g. names of religions).
\item Some DATEs become TIME (e.g. ``earlier tonight'') or EVENT (e.g. ``9/11''), some EVENTs become DATE (e.g. ``Christmas'', ``New Year's Eve''), and TIMEs become DATE if they refer to a duration of more than 24 hours.
\item MONEY mentions that don't include a monetary unit become CARDINAL.
\item Some CARDINALs become ORDINAL (e.g. ``eighth'').
\item Various entity types that are obviously incorrect are corrected (e.g. PERSON for ``New York'', GPE for ``Yeltsin'', DATE for ``Windsor'').
\end{itemize}

Note that some of the tokens that are clearly mislabelled are clustered in sentences or sections where there seems to be a complete mismatch between the tokens and labels. For example, there is a section in the training set (lines 936858-941936) that covers over 5000 lines (i.e. tokens and empty lines) where the tokens and labels are completely mis-aligned. There are at least four or five such sections in the training set. This issue seems to affect mainly one of the six genres in OntoNotes: telephone conversation transcripts.

\paragraph{Ambiguity in annotation: } Unclear annotation guidelines resulted in inconsistent annotations for some cases. For example, in ``Columbia disaster'' and ``Columbia shuttle explosion'', ``Columbia'' is tagged a product, and the other tokens are not tagged, although they should perhaps be tagged events. The annotation guidelines mention ``named hurricanes, battles, wars, sport events, attacks'' as examples of EVENT. However, in ``USS Cole bombing'', ``USS Cole'' is marked a product, and bombing is not tagged. ``Pre cold war'', ``Post september 11'' are tagged events including the pre/post modifiers. In the sentence ``Iran-Contra, Watergate, Monica - they all seem to have a problem'', Monica is tagged as a PERSON although the other two are tagged as events. In this case, Monica refers to the event involving the person, and would have to be tagged an event too. Such cases, where the guidelines are not clear on how to address them, were left as is.

Finally, we also encountered tokenization errors. For example, ``BMW'' is tokenized as ``BMW'' in some contexts, but split into the three tokens ``B'', ``M'', and ``W'' in others; ``post-september'', ``post-resignation'', and ``post-Koizumi'' all appear as a single token, whereas other tokens are split at the hyphen. We also noticed a few typos and POS errors. However, as mentioned in Section~\ref{sec:methods}, we did not attempt to find and correct these.  While what we found through our approach is not exhaustive, and there may have been errors in our own corrections, we believe that the overall effect is a less noisy corpus than before. 

\section{Impact on NER models}
\label{sec:results2}
To understand the impact of the annotation errors on NER, we trained and tested three existing NER libraries using both the original and modified datasets, with the standard train/dev/test split. These systems are described below.

\begin{enumerate}
    \item SpaCy~\cite{ines_montani_2021_5764736} uses a transition-based parser along with a transformer model for NER. It supports four pipelines for English. We used the \textit{en\_core\_web\_trf} pipeline with a RoBERTa-base~\cite{liu2019roberta} and used the provided training script with default settings. Spacy's trainer saves two final models, \textit{model-best} and \textit{model-last}, and we report the results with both models. 
    \item Stanza~\cite{qi2020stanza} provides a Bi-LSTM+CRF NER model with word and character embeddings. We used their pre-trained embeddings and trained the model using the provided training script, with 20K steps.  
    \item Flair~\cite{akbik2019flair} also uses a Bi-LSTM+CRF sequence labeller for NER training, along with stacked embeddings layers. We used the embedding stack with GloVe~\cite{pennington2014glove} and flair embeddings~\cite{akbik2018coling}, which is expected to replicate their results on OntoNotes 5.0 dataset for English.
\end{enumerate}

Table~\ref{tab:nermodels} shows the overall NER performance with the above mentioned models, on old and new datasets. 

\begin{table}[h]
    \centering
    \begin{tabular}{|l|l|p{1cm}|l|p{1.5cm}|} \hline
    \textbf{Model} & \textbf{Data} & \textbf{F1} & \textbf{Delta} & \textbf{Err. red.}\\ \hline
    \multirow{2}{*}{spacy-best} & old & 89.7 & \multirow{2}{*}{1.27}& \multirow{2}{*}{12.33\%}  \\
    & new & 90.97 &&\\ \hline
    \multirow{2}{*}{spacy-last} & old & 89.51 & \multirow{2}{*}{1.03} &\multirow{2}{*}{9.82\%} \\
    & new & 90.54 & & \\ \hline
     \multirow{2}{*}{stanza} & old & 85.92 & \multirow{2}{*}{1.6} &\multirow{2}{*}{11.36\%} \\
    & new & 87.52& & \\ \hline
      \multirow{2}{*}{flair} & old & 89.58&\multirow{2}{*}{1.01}  & \multirow{2}{*}{9.69\%} \\
    & new & 90.59& &\\ \hline
    \end{tabular}
    \caption{Impact on overall NER performance: micro-F1, delta, and relative error reduction.}
    \label{tab:nermodels}
\end{table}

\begin{table*}[ht]
    \centering
    \begin{tabular}{|l|l|l|l|l|l|l|l|l|} \hline
    \multirow{2}{*}{\bf Entity type} & \multicolumn{2}{l|}{\bf spacy-best}& \multicolumn{2}{l|}{\bf spacy-last}& \multicolumn{2}{l|}{\bf stanza}& \multicolumn{2}{l|}{\bf flair}\\ \cline{2-9}
    & \bf Delta & \bf Err. red.& \bf Delta & \bf Err. red.& \bf Delta & \bf Err. red.& \bf Delta & \bf Err. red.\\ \hline
    \multicolumn{9}{c}{\bf Names} \\ \hline
    EVENT &-4.77 &-18.76\% &-7.74 &-30.96\% &1.34 &2.42\% & 1.59&5.94\% \\
    FAC &1.09 & 5\%&1.98 &14.03\% &12.4 & 32.02\% & 7.17& 23.14\%\\
    GPE &0.77 &17.26\% &0.23 &5.31\% & -0.07&-1.34\% &0.34 &8.95\% \\
    LANG &-0.51 &-1.94\% &4.1\% & 15.28\%&2.59 &5.7\% & 6.73& 24.22\%\\
    LAW &3.74 &9.83\% & 1.76& 5.43\%&15.16 &27.79\% &5.67 &15.65\% \\
    LOC &2.07 & 9.78\%& 1.71& 6.93\%&4.75 &18.85\% & 4.37& 18.45\%\\
    NORP &0.87 &16.60\% &-0.34 &-7.46\% &6.38 & 48.88\%& 1.89&31.81\% \\
    ORG & 2.58&23.67\% &1.66 &16.24\% &1.95 &12.39\% & 1.35& 11.92\%\\
    PERSON & 2.47& 38.29\%& -0.68& -7.15\%& 1.19&13.85\% &0.39 & 7.3\%\\
    PRODUCT & -6.48&-30.72\% &3.4 &12.47\% &2.16 &5.48\% & -5.3& -20.25\%\\
    WORK\textunderscore{}OF\textunderscore{}ART &15.13 &40.76\% &10.72 &31.57\% & 13.32& 27.19\%& 17.7& 44.46\%\\ \hline
    \multicolumn{9}{c}{\bf Values} \\ \hline
     CARDINAL & 0.46& 3.26\%& -0.99& -7.29\%&0.1 &0.57\% &-0.41 &-2.94\% \\
     DATE &0.52 &4.19\% &1.98 &14.03\% &0.41 &2.48\% &-0.65&-5.27\%  \\
     MONEY & -2.38&-22.56\% &-2.74 &-26.6\% & -2.31&-19.9\% & -3.91&-39.61\%\\
     ORDINAL & -2.14& -15.12\%& 1.02& 5.94\%& -0.04& -0.24\%& 1.33&8.12\%  \\
     PERCENT & 3.07& 32.55\%& -0.68&-7.15\% & 2.75& 25.32\%&1.19 &13.84\% \\
     QUANTITY & 0.27& 1.47& 1.04& 4.52\%& 4.29& 18.09\%& -1.48&-7.01\% \\
     TIME &0 &0 &3.99 &10.87\% &2.46 &5.82\% & 7.13& 18.82\%\\ \hline
     \end{tabular}  
    \caption{Impact on NER performance by entity type: micro-F1 delta and relative error reduction.}
    \label{tab:nerbycat}
\end{table*}

All the models showed some improvement ($1.01\% - 1.6\%$) in overall performance using the new dataset, with \textit{spacy-best} getting the largest error reduction (12.33\%) among them. This shows that the corrections we made were consistent and improved the models' ability to learn from the data.
\newcite{reiss-etal-2020-identifying} report similar improvements with Flair and a slight drop with a BERT model fine-tuned for NER, on their corrected CONLL-03 dataset for English NER. 

To understand the impact of the annotation errors and corrections in detail, Table ~\ref{tab:nerbycat} breaks down the results by entity type, distinguishing named entities from numerical entities, and showing absolute difference and relative error reduction across models for each entity type.

On average, there is an 8.39\% relative improvement in performance per individual entity type, across all models. Performances vary by individual model, though, and across entity types. Among the named types, FAC, LAW, LOC, ORG, and WORK\textunderscore{}OF\textunderscore{}ART got a performance boost across all models, with WORK\textunderscore{}OF\textunderscore{}ART getting over 10 \% absolute boost with all the models. Among numerical types, TIME got a performance boost with all the models. In terms of a performance drop, EVENT (spacy) and PRODUCT (spacy-best and flair) seem to experience the biggest drop among the named types using the new data. They are also among the least represented entity types in the dataset. Among the values, MONEY saw a large performance drop across all four models. Issues with ambiguous annotation of these  entity types (discussed in Section~\ref{sec:results1}) could have had a role to play in this drop.

Overall, although the annotation experiment changed less than 1.5\% of the tokens overall across train/dev/test sets, it resulted in some overall performance increase across four NER models ($\sim 1.5\%$), and individual NE type performances changed dramatically in some cases with the new dataset (e.g., close to 50\% relative improvement for NORP using Stanza). Thus, we can conclude that the annotation errors we detected and corrected had a strong impact on NER model performance. 

\section{Conclusion}
\label{sec:concl}
We proposed a semi-automatic approach for identifying errors in OntoNotes 5.0 English NER dataset, and corrected them manually. The corrections primarily focused on discrepancies between the guidelines and annotations, and on inconsistency in annotating some of the tokens. Our approach corrected close to 8\% of entity mentions (across all the 18 entity types in the tag set) in each of the standard train/dev/test partitions of the corpus, changing the annotations in close to 10\% of the sentences in the corpus. During this process, we observed several discrepancies between the annotation guidelines and the performed annotations, along with some inconsistent annotations. We also discussed ambiguities in the guidelines for some of the entity types. 

We have shown that these issues clearly impact the NER model performance. In our experiments with training NER models using the original and corrected datasets, models trained on the corrected dataset performed better in terms of overall score, achieving relative performance improvements close to 10\% in all cases. This also resulted in huge performance improvement for some of the previously under-performing entity types with low representation in the dataset. 

\subsection{Outlook}

We do not consider our revision of OntoNotes complete, as in this work, we have focused on precision, not recall -- that is, we started from existing mentions rather than trying to identify mentions that have been overlooked by the annotators. In the future, we plan on investigating whether looking at diff-O tokens (i.e. tokens that are not part of a mention, but usually are in other contexts) could be an efficient way of increasing recall. Future work could also include breaking down results by genre and experimenting with more NER datasets or other sequence labelling tasks (e.g. POS tagging). We also hope to develop a complete, efficient methodology for resource creators to identify and correct annotation errors in sequence labelling tasks. 

\paragraph{Caveat: }  It is important to recognize that our re-annotation may not be perfect: it is not guaranteed to identify all errors and it may have introduced a small quantity of new noise in the attempt to remove old noise. However, our results clearly indicate that the performance of NER models improved with the new version of the dataset, and we hope that this experiment encourages further research in the direction of taking a fresh look at the datasets we use as gold standard, and on developing newer approaches for annotation error detection and correction. 

\paragraph{Language Resources: } Since OntoNotes is licensed via LDC, we cannot release the entire re-annotated corpus publicly, but we can share our corrected NER labels upon request to those who have obtained the license.

\section*{Acknowledgements}
This research was conducted at the National Research Council of Canada, thereby establishing a copyright belonging to the Crown in Right of Canada, that is, to the Government of Canada. 

\section{Bibliographical References}\label{reference}

\bibliographystyle{lrec2022-bib}
\bibliography{lrec2022}


\end{document}